# Multi-agent coordination for data gathering with periodic requests and deliveries

Yaroslav Marchukov and Luis Montano

Instituto de Investigación en Ingeniería de Aragon (I3A),
University of Zaragoza,
C/Mariano Esquillor, s/n, 50018, Zaragoza, Spain
`{yamar, montano}`@unizar.es

## 1 Introduction

In this demo work we develop a method to plan and coordinate a multi-agent team to gather information on demand. The data is periodically requested by a static Operation Center (OC) from changeable goals locations. The mission of the team is to reach these locations, taking measurements and delivering the data to the OC. Due to the limited communication range as well as signal attenuation because of the obstacles, the agents must travel to the OC, to upload the data. The agents can play two roles: ones as workers gathering data, the others as collectors traveling invariant paths for collecting the data of the workers to re-transmit it to the OC. The refreshing time of the delivered information depends on the number of available agents as well as of the scenario. The proposed algorithm finds out the best balance between the number of collectors-workers and the partition of the scenario into working areas in the planning phase, which provides the minimum refreshing time and will be the one executed by the agents.

## 2 Main purpose

Data gathering and delivering by multi-agent systems using collectors have been dealt with in the literature. In [1] the collectors are permanently connected to the central server, they do not need to go to a depot point. In patrolling missions [4], the agents restrict the motion to some locations or areas, or moving the collectors to pre-fixed rendezvous points as in [3]. In this new proposal, we make more flexible the movement of the workers towards their goals, which change in every cycle, and to the collectors. Likewise, the workers establish their meeting area to synchronize with the collectors, dynamically computed, according to the amount of data to share.

## 3 Demonstration

This demo work describes in an illustrative manner the operation of our algorithm proposed in [2], using a team of 20 agents. All the agents are initially located near to the OC, having connectivity with it. The algorithm proceeds in two steps:

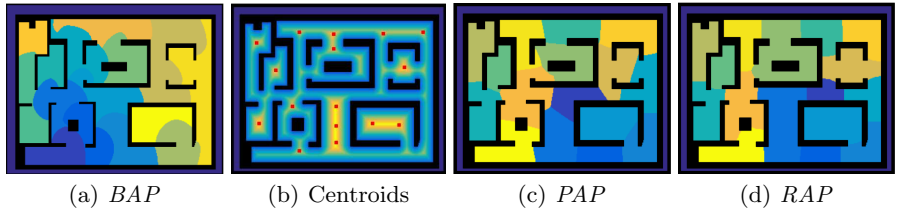

(a) *BAP*  (b) Centroids  (c) *PAP*  (d) *RAP*

**Fig. 1.** Segmentation procedures. (b) depicts the centroids initialization for *PAP-RAP*

1. Planning the deployment: 1) splitting the scenario into working areas for the workers, finding out the best balance of collectors-workers, Fig.1; 2) computing the trajectories of the collectors and associating the workers to share their data, Fig.2. It is a decentralized algorithm that can be executed by the OC or by some of the agents. It is executed offline, previously to deploy the team. The information of the deployment mission is shared between the agents. The first batch of goals are assigned to the workers.
2. Executing the best plan: the trajectories of the workers are planed and executed to visit the goals of their segments, to synchronize with the OC or with a collector in movement, Fig.4(a). They start the mission going to their respective segments. Then, the plan is executed online by the workers when they visit their assigned goals each gathering cycle.

Fast Marching Method (FMM) [5] is used in different parts of the algorithm, solving the scenario segmentation and path planning for workers and collectors.

### 3.1 Planning the deployment

The best ratio collectors-workers is computed for the scenario. The algorithm plans the mission in two steps for each collectors-workers balance.

**Scenario Segmentation** First, the scenario is divided into $N_w$ working areas for the same number of workers. The algorithm tests different partitions using three different segmentation algorithms based on FMM method, depicted in Fig.1: Balanced Area Partition (*BAP*), Polygonal Area Partition (*PAP*), and Room-like Area Partition (*RAP*). *BAP* method uses FMM, extending $N_w$ wavefront until cover all the free space. *PAP* and *RAP* routines consist of two phases: initialization of $N_w$ centroids in areas distant from obstacles, Fig.1(b), and then moving these centroids, balancing the distances between them. In the case of *RAP* the wavefront is expanded in such a way that the resulting segments fit better the shape of the rooms.

**Collectors' paths and associations with workers** The collectors paths and the associations to communicate with the workers are depicted in Fig.2. Firstly, the algorithm obtains the adjacency graph for the centroids of the segments, Fig.2(a). Secondly, an iterative procedure groups the working segments to compute the segments of the collectors for workers-collectors association, taking into account the shape of the worker's segments, Fig.2(b). Third, the paths of the collectors from the OC are obtained, balancing the time of the collectors and of the workers estimated trajectories, Fig.2(c).

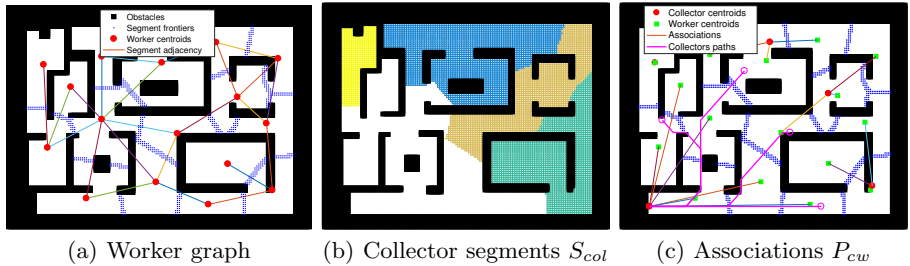

(a) Worker graph  (b) Collector segments $S_{col}$  (c) Associations $P_{cw}$

**Fig. 2.** Collectors' computation 20 agents, 4 collectors and 16 workers, using $PAP$.

### 3.2 Executing the plan

The time devoted by each worker in its associated segment during the execution of the plan depends on the goals to be visited and on the cycle time of its collector. That time is computed to maximize the number of goals to be delivered to the collector or to the OC in the current cycle, Fig.4(a). The workers' trajectories and the area in which they have to synchronize with the collector in movement are computed every cycle according to that objective.

### 3.3 Selecting the best solution

The method selects in the planning phase the best configuration represented by the kind of segmentation ($BAP,PAP,RAP$) and the ratio collectors-workers by means of a Utility function $U$. It balances the number of delivered goals and their refreshing time at the OC. It is computed as $U = \alpha(1-T_{refresh})+\beta N_{goals}$, being $T_{refresh}$ and $N_{goals}$ normalized values. The weighting factors have been set to $\alpha = \beta = 0.5$, giving the same priority to both terms. The refreshing times Fig.3(a), the delivered goals Fig.3(b), and the utilities combining both Fig.3(c), have been computed for $N = 20$ agents and for 0 to 8 collectors. In the planning phase, $U$ is estimated to obtain the best configuration to be executed afterwards. In the figures bands between minimum and maximum values are also shown for several real executions, in order to analyze how well the estimated values had been computed. Figure 4(a) depicts the evolution of the trajectories and synchronization of a worker and a collector during the mission. A snapshot of the execution for the best obtained plan is represented in Fig.4(b) and a simulation can be found in the link [1].

## 4 Conclusions

This work allows to conclude that using some agents in the role of collectors for uploading the information at OC is more efficient that directly moving all the agents to the OC, considering the balance between the refreshing time and the number of delivered goals. The best number of collectors is found within a range 1-4, being the $BAP$ and $PAP$ segmentation methods the best ones depending on the scenario characteristics (see [2] for details). The Room-like partition $RAP$, which is extensively used for SLAM problems, appears as the worst solution for solving the stated deployment problem.

---

[1] http://robots.unizar.es/data/videos/paams19yamar/demo.mp4

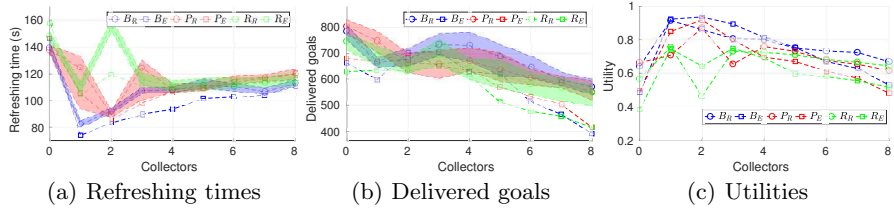

(a) Refreshing times  (b) Delivered goals  (c) Utilities

**Fig. 3.** Values estimated by the planner ($E$) and in the real executions ($R$). The letters $B, P, R$ in the legends refer to $BAP, PAP, RAP$ methods respectively.

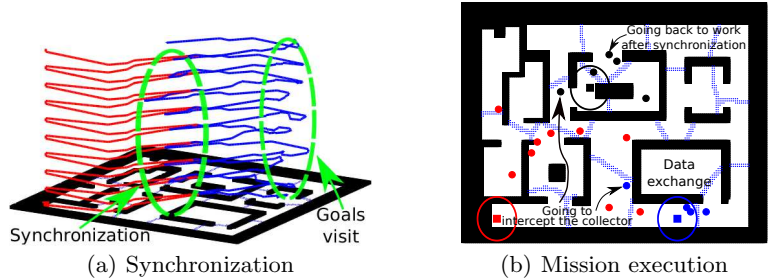

(a) Synchronization  (b) Mission execution

**Fig. 4.** Mission execution. In (a), blue and red lines are a worker and its collector trajectories, respectively. (b) shows a snapshot of a mission execution for $PAP$ and 2 collectors (blue and black squares), the red square OC, and the circular workers associated to each of them (the same colour as the corresponding collector).

## Acknowledgments

This research has been funded by project DPI2016-76676-R-AEI/FEDER-UE and by research grant BES-2013-067405 of MINECO-FEDER, and by project Grupo DGA-T45-17R/FSE

## References

1. Guo, M., Zavlanos, M.M.: Distributed data gathering with buffer constraints and intermittent communication. In: IEEE Int. Conf. on Robotics and Automation (ICRA). pp. 279–284 (May 2017)
2. Marchukov, Y., Montano, L.: Multi-agent coordination for on-demand data gathering with periodic information upload. In: 17th Int. Conf. on Practical Applications of Agents and Multi-Agent Systems (26-28 June 2019)
3. Meghjani, M., Manjanna, S., Dudek, G.: Fast and efficient rendezvous in street networks. In: IEEE/RSJ Int. Conf. on Intelligent Robots and Systems (IROS). pp. 1887–1893 (Oct 2016)
4. Portugal, D., Rocha, R.: Msp algorithm: Multi-robot patrolling based on territory allocation using balanced graph partitioning. In: Proceedings of the 2010 ACM Symposium on Applied Computing. pp. 1271–1276. SAC '10, ACM (2010)
5. Sethian, J.A.: A fast marching level set method for monotonically advancing fronts. Proceedings of the National Academy of Sciences of USA **93**(4), 1591–1595 (Dec 1996)